# Exploring Federated Deep Learning for Standardising Naming Conventions in Radiotherapy Data

Ali Haidar[1] (a.haidar@unsw.edu.au), Daniel Al Mouiee[1,2,3] (d.almouiee@unsw.edu.au), Farhannah Aly[1,2,3] (f.aly@unsw.edu.au), David Thwaites[4,5] (david.thwaites@sydney.edu.au), Lois Holloway[1,2,3,4] (lois.holloway@unsw.edu.au)

[1]South Western Sydney Clinical School, University of New South Wales, Liverpool, Australia
[2]Ingham Institute for Applied Medical Research, Liverpool, Australia
[3]Liverpool and Macarthur Cancer Therapy Centres, Liverpool, Australia
[4]School of Physics, Institute of Medical Physics, University of Sydney, Camperdown, Australia
[5]Radiotherapy Research Group, Leeds Institute of Medical Research, St James's Hospital and University of Leeds, Leeds, UK

**Corresponding author**: emails: ali.hdrv@outlook.com | a.haidar@unsw.edu.au

*Abstract*— Standardising structure volume names in radiotherapy (RT) data is necessary to enable data mining and analyses, especially across multi-institutional centres. This process is time and resource intensive, which highlights the need for new automated and efficient approaches to handle the task. Several machine learning-based methods have been proposed and evaluated to standardise nomenclature. However, no studies have considered that RT patient records are distributed across multiple data centres with restrictions on being transferred and aggregated in one location. This paper introduces a method that emulates real-world environments to establish standardised nomenclature. This is achieved by integrating decentralised real-time data and federated deep learning. A multimodal deep artificial neural network was proposed to standardise RT data in federated settings. A dataset of lung cancer patients treated with RT was collected from The Cancer Imaging Archive (TCIA). Three types of possible attributes were extracted from the structures to train the deep learning models: tabular (1D), visual (2D), and volumetric (3D). Simulated experiments were carried out to train the models across several scenarios including multiple data centres (3, 5, and 7 data centres), input modalities (tabular-visual, tabular-volumes), and aggregation strategies (FedAvg, FedAdam, FedOpt, and FedYogi). The models were compared against models developed with single modalities (tabular, visual, volumetric) in federated settings, in addition to models trained in centralised settings. Categorical classification accuracy was calculated on hold-out samples to inform the models performance. Our results highlight the need for fusing multiple modalities when training such models, with better performance reported with tabular-volumetric models. In addition, we report comparable classification accuracy compared to models built in centralised settings. This demonstrates the suitability of federated learning for handling the standardization task. Additional ablation analyses showed that the total number of samples in the data centres and the number of data centres highly affects the training process and should be carefully considered when building standardisation models.

## 1. INTRODUCTION

Radiotherapy (RT) data from routine clinical practice for cancer treatment provide information on what happens to patients in the real world. Learning from such data provides the potential to enhance understanding and optimise decision making for individual cancer patients. Applying data mining on large datasets has been demonstrated to improve knowledge in clinical challenges [1]. Indeed, machine learning algorithms have been widely used in RT, with applications in outcome prediction, survival analyses, and segmentation [2-8].

However, transferring de-identified records between institutions has ethical (privacy) considerations and is a complex task. To overcome such limitations, federated systems have been introduced such as EuroCAT and the Australian Computer-Assisted Theragnostics (AusCAT) networks [9, 10]. The EuroCAT network was initially established across three European countries [9]. The AusCAT network was established across national and international radiation oncology departments, to enable data mining and learning from clinical practice datasets including radiotherapy [10]. Learning across such networks occurs without the data leaving the individual centres, known as distributed or federated learning. This approach enables extracting trends and patterns from retrospective clinical data, with the potential for this evidence to be used to supplement clinical trial datasets, which are smaller and with closely defined eligibility criteria.

Federated networks enable exploring various research questions in radiation oncology, such as dosimetry analyses in lung and breast cancer, predictive outcome modelling and model validation in lung cancer and head-and-neck cancer, auto-segmentation, guidelines, and compliance analyses [11-14]. All these research questions require collection and standardization of RT data, which includes three-dimensional data such as structure names often stored in Digital Imaging and Communications in Medicine (DICOM) formats. Previous work has shown that 80% of work with medical data is the process of cleaning, curating and preparing medical records [15], required before any model development can start.

A RT plan consists of structure volumes contoured by clinicians, or sometimes automatically, that represent target volumes (TVs), organs-at-risk (OARs), and optimisation structures. The TVs are the regions identified as requiring radiation treatment. OARs are organs lying close to the high radiation dose target (tumour) region and which are potentially at most risk of unwanted radiation effects and which must be considered carefully in planning a RT treatment. The optimisation structures are parameters introduced during the treatment planning process to optimise the RT plan. The clinicians optimise the dose to be received by each volume to ensure the intended dose is delivered to the target as well as avoid damaging healthy tissues. The name of each volume is described in the form of text, which has until recently not been standardised across patients and data centres. Although recent standardising naming recommendations have been defined [16], these are not always followed. An example of OARs in a lung RT plan is shown in Figure 1.

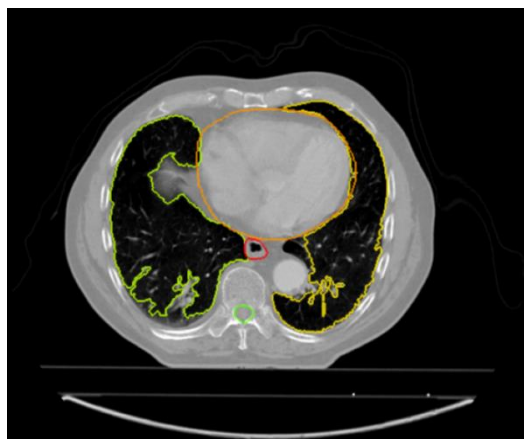

Figure 1 n example of OARs outlined on a CT scan for RT planning of a lung cancer patient (scan viewed looking towards the patient's head): left-lung (yellow), right-lung (light green), heart (orange), esophagus (red), spinal cord (green).

Specifically, inconsistency can occur in the naming of patients' TV and OAR volumes in RT plans. This might be due to several reasons, such as the lack of specific templates/protocols for contouring and naming the volumes, variations in time, or between clinicians or institutions in contouring different patients' plans, human error, etc. To utilise retrospective datasets, clinicians and researchers usually interact to handle variations. Clinicians try to group the variations under common names, and researchers implement the tools required to set up the rules that simulate the clinicians' decisions. This process is time consuming, which increases rapidly in two scenarios: expanding the number of centres and adding new patient records to the data centre. Hence, frequent meetings with the clinicians are required to discuss and update the structure standardization processes. Consequently, there is a need for new systems to standardise the process of RT imaging data structures and automate their extraction.

Machine learning based methods have been deployed for standardising naming conventions in RT data [17-22]. Rozario et al proposed a neural networks based approach to standardise nomenclature in prostate and head-and-neck cancer patients [23]. The proposed architecture consisted of three convolutional blocks with two fully connected layers, where 2D images were used for training the model. Yang et al proposed a 3D convolutional neural network (CNN) with adaptive sampling and adaptive cropping (ASAC) that utilised volumes to standardise 28 OARs in head-and-neck cancer patients [21, 24]. Sleeman et al investigated multiple algorithms for standardising nomenclature to the American Association of Physicists in Medicine(AAPM) TG-263 protocol naming conventions for prostate and lung cancer patients [20]. Bitmap images were used, where features were extracted to train gradient boosting decision trees (GBT), support vector machines (SVM), multi-layer perceptron (MLP), Naïve Bayes (NB) and random forest (RF) classifiers. The datasets belonged to multiple centres and were integrated to develop the model. Other approaches have used similar datasets, e.g., utilising text features and a neural network [25], or integrating geometric and text features to train a RF to standardise nomenclature [18]. A deep learning approach was deployed to standardise volumes in a large prostate cancer dataset, where the ResNetV2 was used to build the classification model [26]. Haidar et al used neural networks and multimodal learning to standardise TV and OARs in a breast cancer dataset [19].

Various types of input modalities have been incorporated to perform the classification task across different cancer sites: lung, breast, prostate, and head-and-neck. High performance has been reported in standardisation applications, with most of the developed models revealing classification accuracies higher than 90%. This performance is expected in this type of application, taking into consideration the ability to establish rich representations of the volumes and their anatomical variations. However, all the previously reported studies in the literature used models that were developed in centralised settings, which would still be limited within distributed networks. It is not always possible to aggregate RT data in one location taking into consideration data privacy and protocols. To facilitate learning from distributed data to standardise nomenclature, this paper proposes and evaluates a federated deep learning-based approach for automating the process of standardising retrospective RT imaging data, while keeping the records at the data centres. This paper is organised as follows: Section 2 reports the proposed method and provides an overview of the steps involved for preparing the dataset. Section 3 describes experimental setup, results and analyses, and key limitations. Section 4 derives a conclusion and highlights future work.

## 2. MATERIALS AND METHODS

In this section, we outline the materials and methods used to standardise nomenclature using federated deep learning. First, the data collection and feature extraction methods are described. Second, the proposed network architecture is explained. Then, the proposed approach that utilised federated deep learning is presented.

### 2.1 Data

#### 2.1.1 Data Collection

A public dataset that consisted of 422 lung cancer patients treated with RT was collected from The Cancer Imaging Archive (TCIA) [27]. The dataset has been used in survival modelling, where the RT plan gross tumour volumes (GTVs) were used in outcome prediction and survival analyses [28]. The dataset contained the computed tomography (CT) images, and the DICOM RT structure set (RTSTRUCTS) that contain the structures contoured by the clinicians. The dataset was collected in DICOM format before being pushed to an Orthanc server, which is an open-source DICOM picture archiving and communication system (PACS) [29]. The Orthanc server enables indexation of DICOM data and provides a web server for data visualisation. PDCP, a tool that enables processing patients records into machine learning usable formats, was used to extract the volumes stored in the Orthanc server [30].

#### 2.1.2 Feature Extraction and Data Partitioning

A set of seven classes was identified for use in this work, that constituted one TV and six OARs: GTV-1, Spinal-Cord, Esophagus, Lung-Left, Lung-Right, Heart, and Lungs-Total. Additional volumes were out of scope for this work (e.g., GTV-2) and were removed from the patients' RT plans. Three types of attributes/features were extracted from contoured volume belonging to any of the selected seven classes:

- **Tabular Features:** are defined as 1D information holding numerical representations about the position of the volume in the RT plan compared to a reference point (0,0,0), in addition to details such as the number of voxels in a volume. Nine features were extracted to represent each structure position and size. More comprehensive insights into the generation process of these features can be found in Haidar et al [19].

- **Visual Features:** are defined as 2D information representing the slices with the highest number of contoured pixels across one of the planes in the volumes (axial plane). These features are also known as central slices and have been used in outcome prediction tasks in the literature, where slices were extracted from contouring masks to predict patient outcomes in a head-and-neck dataset [31]. In the current work, the contour mask was laid on the CT volume, before extracting the central slice. The central slice extraction used information on voxel Hounsfield Units (HUs), which represent the relative (greyscale) radiodensities of the voxels in the CT images. Since the slice values varied between structures, we applied minimal and maximal thresholds for standardisation purposes. The central slices were resized to 64*64 arrays to train the deep neural networks.

- **Volumetric Features:** are defined as 3D data representing the structures volumes in the RT plan. All the extracted volumes were resized to 64*64*32 arrays to train the deep neural networks. Since the HUs for the voxels of a given volume vary between structures, minimal and maximal thresholds were applied while extracting the volumes to standardise the voxels values.

The collected dataset was divided into two subsets: model development and model testing. The records collected from 372 patients were used for development and validation, and the records collected from the last 50 patients were utilised for testing (hold-out sample). Table 1 reports the number of classes in each subset.

Table 1 Number of samples in each subsets (training +validation, test).

| Class | Number of Samples | Training and Validation | Test |
|---|---|---|---|
| GTV-1 | 421 | 371 | 50 |
| Spinal-Cord | 411 | 361 | 50 |
| Esophagus | 355 | 307 | 48 |
| Lung-Left | 312 | 289 | 23 |
| Lung-Right | 312 | 289 | 23 |
| Heart | 127 | 78 | 49 |
| Lung-Total | 97 | 70 | 27 |
| **Total** | 2035 | 1765 | 270 |

## 2.2 Artificial Neural Networks and Multimodal Learning

The proposed approach incorporated a multimodal learning paradigm to build the standardisation model. Three levels of multimodal learning can be used to fuse modalities: input-level, layer-level, and decision-level [32]. We propose a layer-level approach, where the modalities are concatenated inside the neural network, with a single last layer responsible for handling the predictions, as shown in Figure *2*.

The network architecture consisted of fully connected (FC), convolutional (conv), pooling, dropout, and batch normalization (BN) layers. The FC blocks were used to train the tabular features. The convolutional blocks were utilised to train imaging features or volumes. Within this study, we aimed to minimise the number of weights to be transmitted between the orchestrater (next section) and the data centres while training, hereafter minimising communication cost. For this reason, we investigated small convolutional blocks (three layers of conv, BN, pooling, dropout) to train the central slices and volumes.

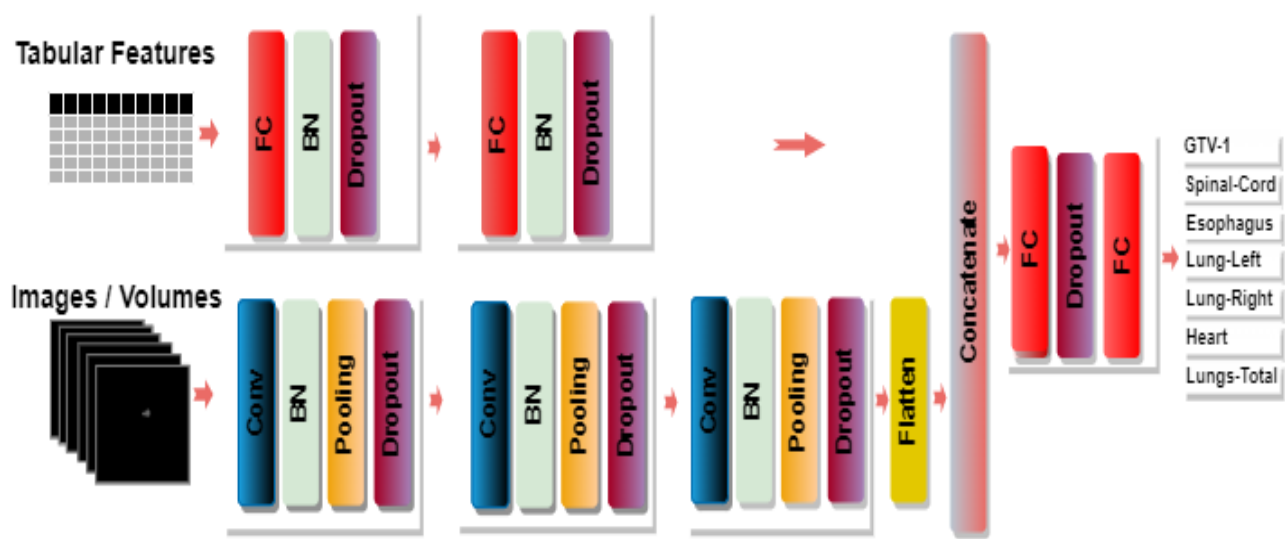

Figure 2 Proposed network architecture.

**2.3 Federated Learning**

Federated learning (FL) is a mechanism for training machine learning algorithms while keeping the datasets distributed at the client data centres [33, 34]. It consists of a server (orchestrator) that manages the training process, and a set of clients that host the datasets and training rounds. The central server is responsible for handling various tasks: setting up connections to the centres, mechanisms for handling requests/responses between the orchestrator and data centres, aggregating model updates following training rounds at each centre, evaluation, and monitoring of the training process. The clients contain the datasets used for training and evaluation.

Several open-source platforms have been introduced in recent years to enable federated learning [35-37]. TensorFlow Federated (TFF) is a framework for deep learning and other computations developed by Google [38]; Pysfyt is a framework that extended existing tools such as TensorFlow [38] and Keras [39] to enable private and secure deep learning [36]. FATE is an open-source framework that enables training deep learning algorithms as well as others (such as tree-based algorithms); FedML is another open-source tool that enables federated learning [40]; Flower is an open source framework that facilitates training deep learning algorithms across multiple devices holding local data samples [37]. In addition, flower enables federated learning evaluation, while interacting with its clients using the Google Remote Procedure Calls (gRPC) protocol. As reported by its developers, the flower framework contains tools that enable setting up a central server, aggregating model updates, evaluating models, handling requests, and monitoring the training process. From our perspective, we found that flower was easy to incorporate in our task and was selected for use in this study.

The proposed approach is shown in Figure 3. The pseudocode of the federated learning training is shown in Algorithm 1. Initially, (a) the datasets are created and pre-processed at the clients; (b) the training process starts with the central server sending configuration files to each client; (c) the parameters of the model are then shared with the clients allocated for training. The model is trained on each client's dataset (typically for 1 epoch), before sending the updated weights back to the server. The updates are then grouped based on the selected aggregation method. For deep learning, the connection weights between the neural processing units (and some other parameters) are updated following training rounds across the distributed datasets before being fused/aggregated in the server. (d) the aggregated weights are then shared with clients allocated for evaluation. The process is repeated until reaching a stopping criterion (e.g., training rounds have been reached).

Different methods have been implemented to enable aggregating the weights after each training round at the clients. FedAvg is the simplest aggregation method that takes the average of the weights of the connections in the neural network layers and neurons [41]. New approaches have been introduced later to enhance model performance and to enable adaptation such as FedOpt, FedAdam, and FedYogi [42]. With adaptation, the learning rate of the algorithm is updated while training the algorithm, which has shown better performance in centralised settings.

The federated learning process in RT standardisation was defined as a cross-silo [43], taking into consideration that all the clients are required to participate in training rounds, and the number of clients is relatively small compared to cross-device federated learning settings. To evaluate the model while training, we proposed federated evaluation compared to central evaluation, where all the centres were used in assessing the aggregated model performance. In addition to, validation samples were allocated within the centres to avoid model overfitting while training.

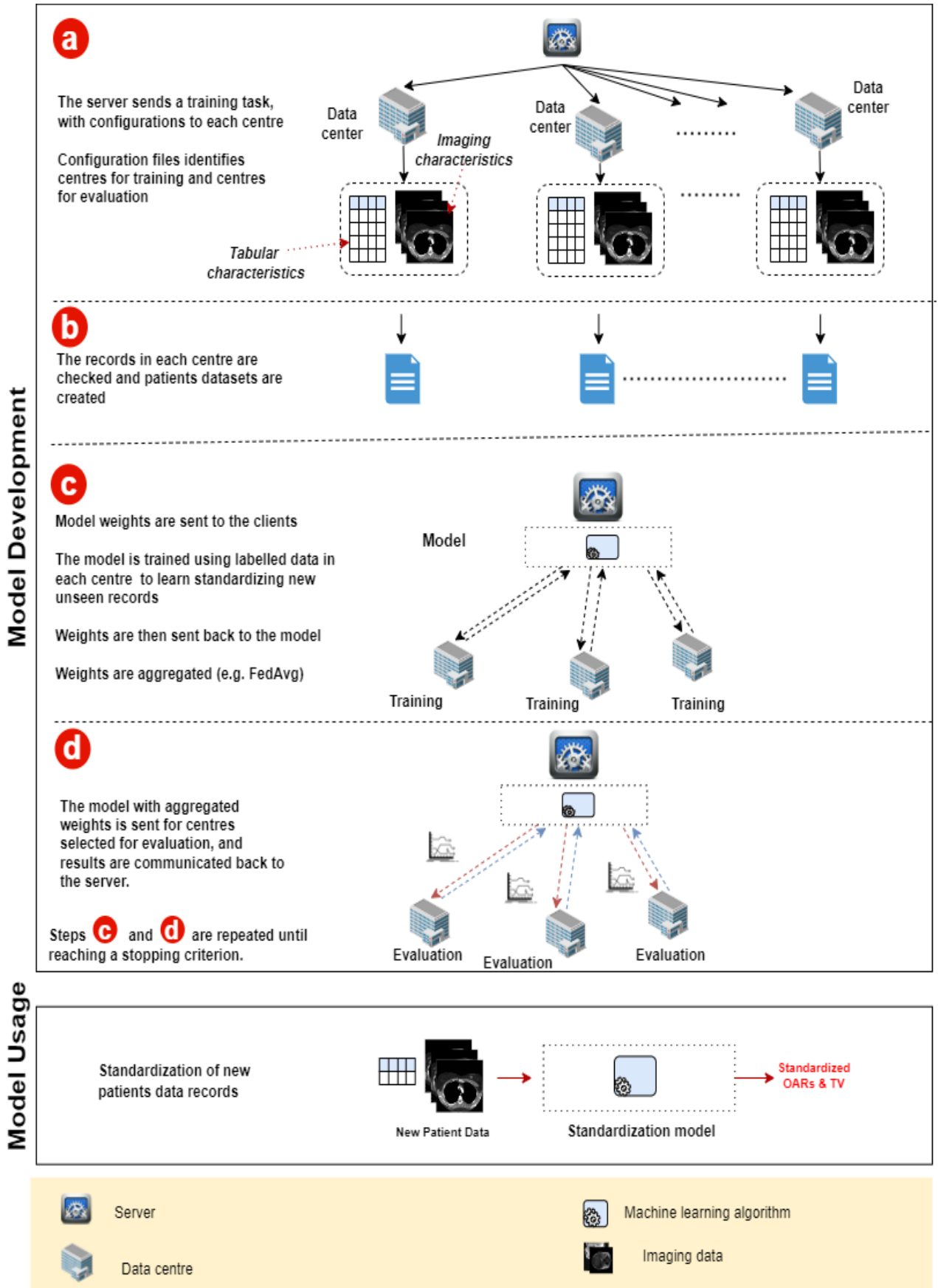

Figure 3 An overview of the proposed approach for standardizing nomenclature in RT data.

| Algorithm 1 | Federated Learning for RT Data Standardisation |
|---|---|
| **Output** | ANN *weights* |
| **Setup** | *central-server (Orchestrator) and datacentres* |
| *Orchestrator* | **Input:** training-centres list, evaluation-centres list, fedStrategy, training rounds<br>create the ANN architecture, allocate initial *weights* to *model*<br>initialize the fedStrategy with initial *weights*<br>**while** training-centres required to start training not satisfied **do** wait |
| *Datacentres* | **Input**: ($training_c$, $validation_c$) **for** $c$ **in** $training + evaluation$ data<br>**for** c **in** $training + evaluation$ centres **do**<br>    prepare $training_c$ and $validation_c$ samples<br>    create the neural network architecture *model*<br>    initiate a request<br>    connect to the *central-server* |
| **Model Development** | |
| | **for** training round **in** training rounds **do** |
| *Orchestrator* | check list of training-centres<br>**for** $tc$ **in** training-centres **do**<br>    initiate a request<br>    add a training task<br>    add the *weights*<br>    send the request to centre<br>**while** not training-centres responses ready **and** no timeout **do** wait |
| *Training datacentres* | **for** $tc$ **in** training-centres **do**<br>    unpack the request<br>    assign *weights* to model<br>    train *model* for one epoch<br>    initiate a response<br>    add the updated weights to the response<br>    send the response to the orchestrator |
| *Orchestrator* | **for** response **in** responses **do**<br>    unpack the response<br>    get the weights and add to list-of-weights<br>*weights* ← aggregate list-of-weights based on fedStrategy<br>**for** $ec$ **in** evaluation-centres **do**<br>    initiate a request<br>    add an evaluation request<br>    add the new aggregated *weights*<br>    send an evaluation request to the centre<br>**while** not evaluation-centres responses ready **and** no timeout **do** wait |
| *Evaluation datacentres* | **for** $ec$ **in** evaluation-centres **do**<br>    unpack the *weights*<br>    assign *weights* to model<br>    evaluate on the validation samples<br>    initiate a response<br>    add the evaluation metrics to response<br>    send response to the central server |
| *Orchestrator* | **for** metric **in** metrics **do** aggregate metrics<br>save the new *weights* |

## 3. EXPERIMENTS AND RESULTS

### 3.1 Evaluation Metrics, Case Studies, and Experimental Setup

The study aimed to investigate the applicability of federated learning based methods in RT data standardisation. To evaluate the performance of the models, categorical accuracy was computed across the hold-out samples. Categorical accuracy reports the percentage of the correctly classified samples compared to the total number of samples. Multiple case studies/scenarios were introduced by varying the number of clients, input modalities and aggregation strategies:

- Number of clients: 3, 5, and 7 centres.
- Input modalities: tabular-visual, tabular-volumes
- Aggregation strategies: FedAvg, FedOpt, FedYogi, and FedAdam

For comparison purposes, feed forward neural networks (FFNNs) and CNNs were implemented to train each single modality (tabular, visual, volumes) separately, with the same parameters (clients and aggregation methods). The FFNN architecture was identical to the fully connected blocks that handled the tabular data in the proposed multimodal network. The CNN architecture was the same as the imaging architecture in the multimodal network, in addition to a fully connected layer before the last layer.

Furthermore, the datasets (training and validation) used in federated settings were combined and trained on centralised settings with the same architectures. To enable better comparison and generalizability analyses, each experiment was repeated three times, with the mean categorical accuracy taken to report performance.

As shown in Table 1, the dataset consisted of 372 patients for training and validation. Distributing the patients roughly equally between the centres, the number of patients per centre varied based on the number of centres, with approximately 124, 74, and 53 patients each for the three different scenarios (3, 5, and 7 data centres). The number of classes varied in each data centre, as not all the patients would have the same number of contoured structures. The same experimental setup was employed with all the training scenarios, with the Adam optimizer and a learning rate of 0.01 [44]. The batch size was selected to 16 and number of epochs to 100, with categorical cross entropy as the loss function.

### 3.2 Experimental Results

Table 2 and 3 report the average categorical accuracy and its corresponding standard deviation on the hold-out sample (50 patients) in federated and centralised settings respectively. In 10 out of the 12 experiments, using tabular-volumes based neural networks derived better models compared to tabular-visual neural networks. This was expected as the volumes contain richer representations about the structure. However, training models with combined tabular and visual modalities was faster compared to models with combined tabular and volumes modalities. The largest difference in performance between the multimodal and single networks was with tabular data, followed by images, then volumes. This was also expected for the same aforementioned reason. As mentioned previously, high performance is required to enable standardisation in this kind of application and so using such tabular features models only is not applicable.

Table 2 Models performance over the hold-out sample in federated settings based on number of centres, aggregation method, and data types.

| Centres | Aggregation | Single Modalities | | | Multiple Modalities | |
| --- | --- | --- | --- | --- | --- | --- |
| | | Tabular | Visual | Volumes | Tabular – Visual (proposed) | Tabular – Volumes (proposed) |
| 3 | FedAvg | 87.28 (2.26) | 91.11 (1.28) | 98.27 (0.57) | 93.46 (0.77) | 98.03 (0.43) |
| | FedOpt | 88.89 (2.96) | 91.11 (1.34) | 98.15 (0.37) | 93.58 (0.77) | 98.52 (0.37) |
| | FedYogi | 86.42 (2.17) | 91.11 (0.98) | 98.03 (0.43) | 92.84 (0.43) | 98.52 (0.00) |
| | FedAdam | 88.39 (4.45) | 92.35 (0.21) | 98.02 (0.21) | 93.21 (1.07) | 96.30 (4.51) |
| 5 | FedAvg | 87.9 (2.38) | 92.1 (0.43) | 98.27 (1.4) | 93.46 (0.43) | 97.41 (2.31) |
| | FedOpt | 89.75 (0.77) | 91.73 (0.43) | 97.04 (2.59) | 93.33 (0.37) | 97.65 (1.19) |
| | FedYogi | 88.02 (0.93) | 91.36 (0.43) | 97.53 (1.4) | 93.33 (0.37) | 98.52 (0.98) |
| | FedAdam | 88.27 (3.11) | 91.61 (0.77) | 97.90 (0.57) | 93.70 (0.98) | **99.01 (0.57)** |
| 7 | FedAvg | 87.90 (2.23) | 90.37 (0.37) | 97.04 (1.28) | 93.09 (0.93) | 91.98 (4.32) |
| | FedOpt | 88.52 (2.89) | 91.73 (0.77) | 95.80 (3.50) | 93.09 (0.21) | 98.77 (0.57) |
| | FedYogi | 88.27 (2.04) | 92.84 (0.93) | 97.90 (0.21) | 92.84 (0.21) | 98.27 (1.19) |
| | FedAdam | 88.89 (0.74) | 91.85 (1.11) | 98.15 (1.61) | 92.47 (0.43) | 98.40 (1.30) |

In general, using multiple modalities led to better performance. Nevertheless, in the three environments (3, 5, and 7 centres), using FedAvg for aggregation in both multimodal settings revealed models with lower accuracy compared to the volumes-based models. It was observed that FedAvg was not the best option for multimodal learning, and accuracy degrades while increasing the number of data centres.

Comparable accuracy was obtained in federated and centralised settings, although the centralised tabular-volumes model showed better performance in 10 out of the 12 experiments compared to the models built through federated learning. Better federated performance was reported with 5 centres, FedAdam, and tabular-volumes modalities (99.01%). The same performance (federated to centralised) was reported with 7 centres, FedOpt, and tabular-volumes modalities (98.77%). We assume this occurred because training the model via distributed settings enabled further "attention" on classes with lower numbers of records compared to the majority classes. Federated training avoided "attention" on classes with higher numbers of samples, as in centralised settings, which led to higher accuracy. Hence, using a federated setting in RT standardisation showed comparable performance to centralised settings. Increasing the number of centres would not necessarily lead to degradation in performance. Furthermore, the federated aggregation strategy should be carefully chosen to ensure comparable performance, and no strategy has clearly outperformed the others.

Table 3 Models performance over the hold-out sample in centralised settings.

| Data Type | Performance |
| --- | --- |
| Tabular | 92.1 (0.57) |
| Visual | 91.36 (1.19) |
| Volumes | 98.64 (0.43) |
| Tabular – Visual | 93.95 (0.57) |
| Tabular – Volumes | 98.77 (0.21) |

The best performing model revealed a categorical accuracy of 99.63% (5 centres, FedAdam, tabular-volumes). To analyse features contribution on the hold-out sample, we applied t-distributed stochastic neighbour embedding (t-SNE) on three different layers outputs in the trained multimodal neural network. t-SNE is a mechanism for feature reduction that preserves the original information of high dimensional space. t-SNE compares each pair in the dataset, and clusters pairs with similar information. t-SNE has been used to visualise how well a model has learned to divide the classes on each block. Figure 4 reports the output of the t-SNE over each of the three layers: (a) last layer in the fully connected block used to train tabular features (b) last layer in the convolutional block used to train volumes (c) last hidden layer in the trained network.

It was observed that the fully connected block was capable of differentiating some classes (heart and lungs) and has failed with others (GTV-1 and Spinal Cord). The convolutional block had successfully differentiated the classes into different clusters. In the last hidden layer, classes were successfully differentiated, and the distance between the

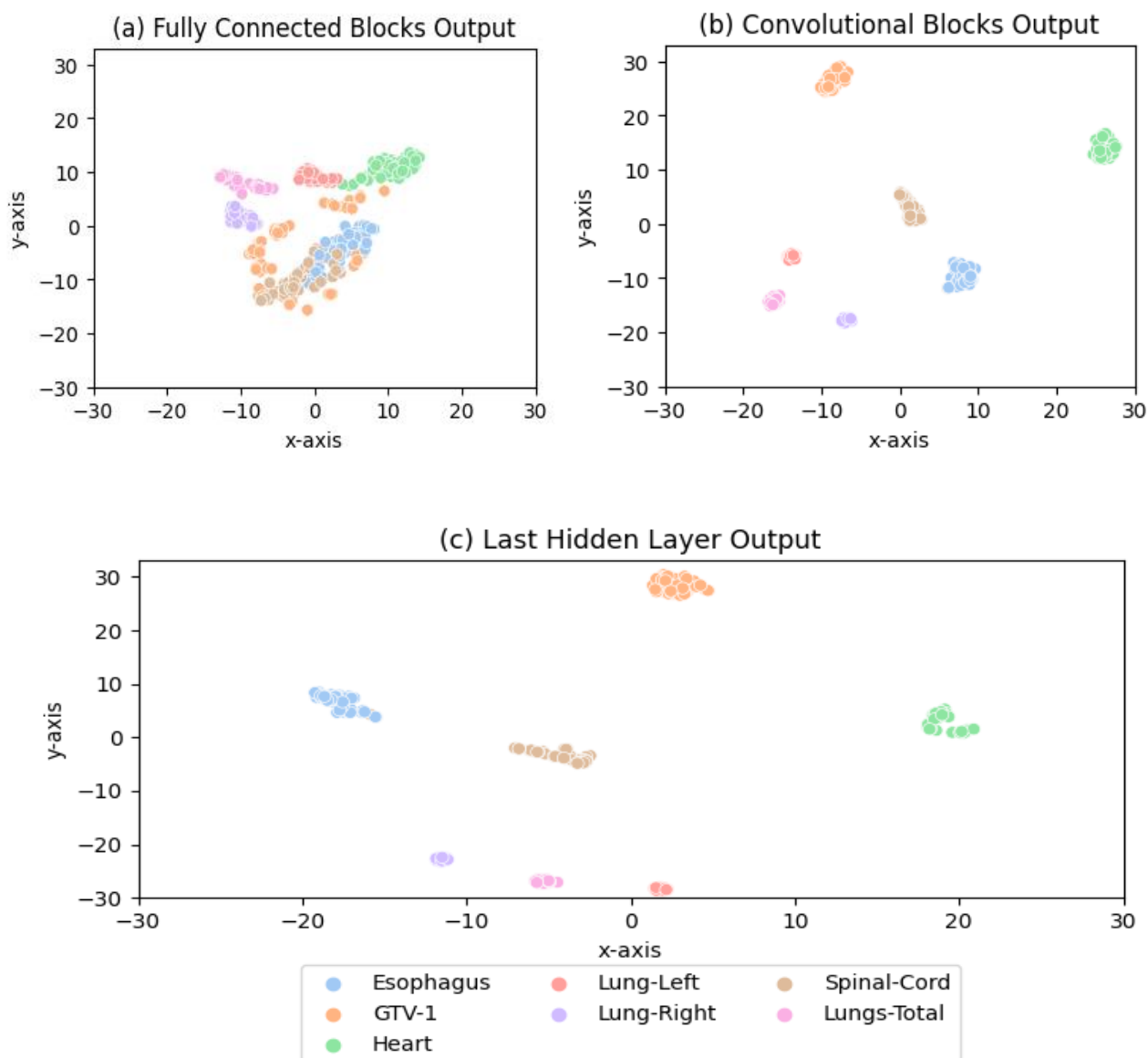

Figure 4 t-SNE on layers outputs in the best performing model: (a) t-SNE on the outputs from the last layer in the fully connected block (b) t-SNE on the outputs from the last layer in the convolutional block (c) t-SNE on the outputs from the last hidden layer in the multimodal network.

clusters was larger compared to the convolutional block. The volumetric features highly contributed to the output of the model, while the tabular features contributed to differentiating some classes. This integration allowed for better models compared to models with single modalities.

**3.3 Ablation Study**

According to Zhang et al, the amount of data represents a key challenge in federated learning [45]. When using machine learning algorithms to standardise RT data, the clinicians and researchers collaborate to manually label the dataset, to enable supervised machine learning or to set up rules to enable weak supervised learning. The number of samples to be used in training a standardisation model is not always clear, especially that it is commonly known that deep learning requires a high number of samples to enable generating distinctive representations [46]. However, in RT standardisation, some classes' features are diverse which might not require a high number of samples to enable classification and standardisation. For example, the lung voxel values are distinctive compared to the heart voxel values, similarly high differences can be identified based on the shape. This explains the high performance usually reported in the literature.

We conducted an ablation study to analyse the effect of the number of samples on the overall performance of the multimodal networks. Two scenarios were introduced, the first scenario was to use only 50% (883 samples approx. of the 7 classes) of the samples belonging to each class in each data centre. The second scenario was to use only 25% (442 samples approx.) of the total number of samples belonging to each class in each data centre. For example, if there were 10 (100%) GTV samples in the first data centre, 5 (50%) GTV samples were selected in that centre with the first scenario experiments and 3 (25%) GTV samples were selected in that centre with the second scenario. It is worth mentioning that the data distribution in the centres was based on the number of patients. Hence, the class distribution was distinct. The same network architecture, experimental setup, aggregation method, and training rounds were used in the ablation experiments with the multimodal networks. Similarly, the same experiments were conducted in centralised settings.

Figure 5 reports the model performance (categorical accuracy) on the same hold-out dataset in each of the four federated aggregation methods, using 100%, 50%, and 25% of the samples representing each class in each centre. In addition, the results in a centralised setting were included for comparison purposes. Comparable performance was obtained when using 50% of the samples in each data centre in federated settings across the three different scenarios (3, 5, and 7 data centres). With 25% of the samples, the models still performed reasonably well in federated settings with 3 centres. However, model performance degraded exponentially when distributing the classes across more centres. It was noticed that the models were not able to generalise with 7 data centres. It was concluded that integrating low numbers of samples with traditional learning will not be effective in RT data standardisation. There is a need to ensure a sufficient number of samples to be labelled in each data centre, no matter the number of data centres. In addition, the aggregation methods are highly dependent on the number of samples in single centres.

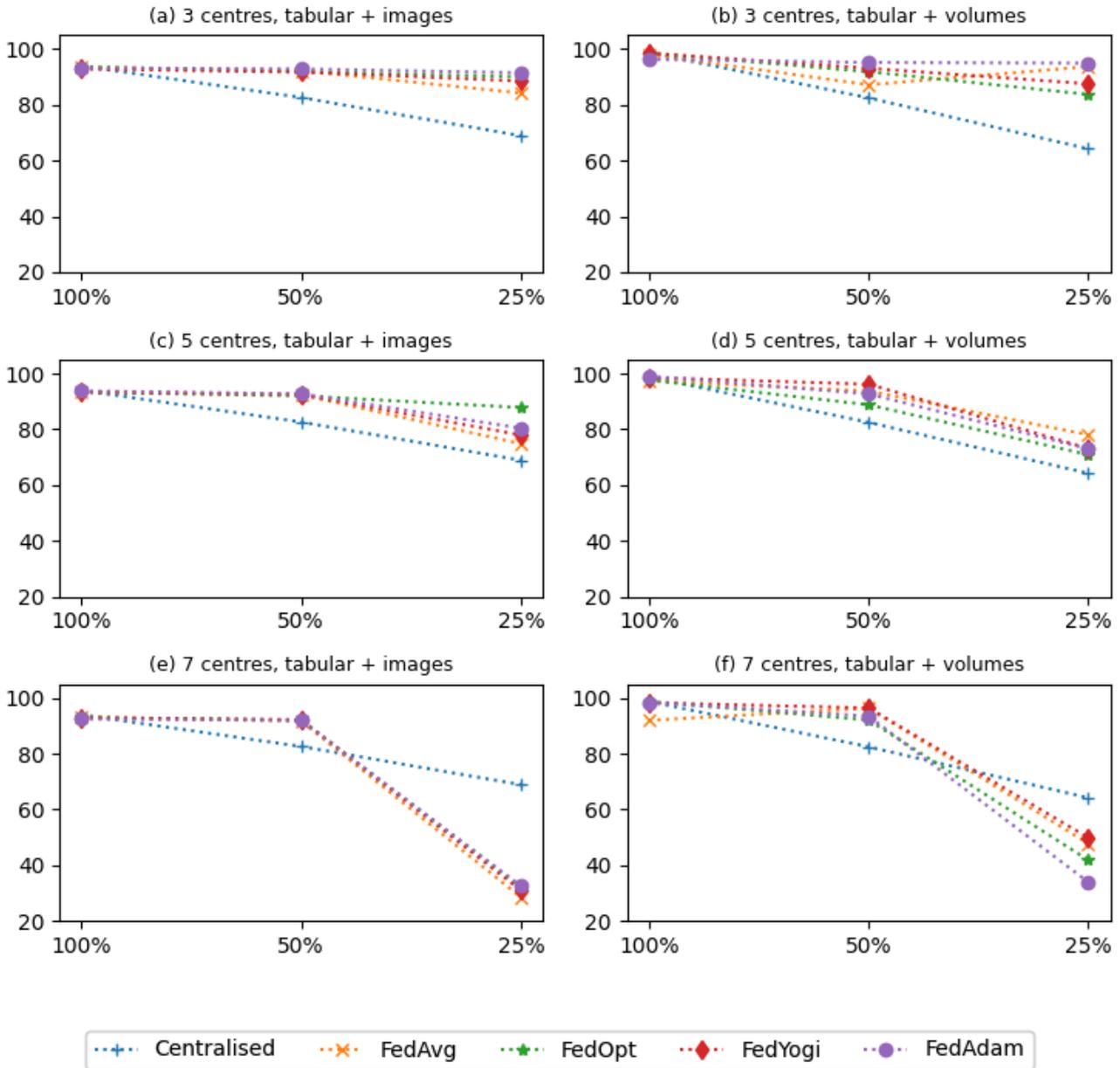

Figure 5 Model performance on each of the multimodal networks in each scenario with 100%, 50%, and 25% of the samples in each data centre.: (a) 3 centres with tabular and visual data; (b) 3 centres with tabular and volume data; (c) 5 centres with tabular and imaging data; (d) 5 centres with tabular and volumes data; (e) 7 centres with tabular and visual data; (f) 7 centres with tabular and volumes data.

### 3.4 Limitations and Future Work

One of the key limitations of this work was the simulation of data centres, although this was effective for comparison purposes. Nevertheless, similar methodology was followed in the literature by using datasets such as CIFAR10 in federated models development and validation [47]. In future work, datasets from multiple data centres will be targeted, such as the head-and-neck cancer datasets collected from four different centres [2]. In addition, seven classes were utilised in this work. To enable deploying such models in real-world scenarios, there is a need to include all the RT plan volumes in the standardisation task as in [19]. In future work, all the TV, OARs, and optimisation structures will be included in training the models.

It was observed that supervised machine learning becomes ineffective with smaller samples distributed across multiple data centres. Few-shot learning will be targeted in future methodology with such scenarios [48]. Last but not least, the models were developed with no augmentation methods. The use of augmentation methods will be targeted in RT standardisation as it was proved to be useful in RT applications [49].

## 4. CONCLUSION

This paper presented and analysed a new approach for standardising nomenclatures for target and organs-at-risk volumes in radiotherapy (RT) data using federated deep learning and multiple feature modalities. The proposed approach was investigated over multiple scenarios by varying the number of centres, number of samples in data centres, aggregation methods, and input features modalities. It was concluded that federated learning can achieve comparable performance to centralised learning for this application, and a model trained via distributed settings can be useful in such scenarios. We analysed the effect of the number of samples on model performance, showing that the data distribution across centres heavily influences the model's performance. Finally, the aggregation method should be carefully chosen in such applications when sufficient data are available. An aggregation method performing well with single modality based models does not imply it will be performing well with multiple modalities.

AVAILABILITY OF DATA AND CODE

The code will be made accessible upon publication of the research paper: https://github.com/ayhaidar/FedRTStandardisation. The dataset is available at the following location: https://wiki.cancerimagingarchive.net/display/Public/NSCLC-Radiomics .